\documentclass[conference,onecolumn]{IEEEtran}

\usepackage{icomma}
\usepackage{soul}
\usepackage{cite}
\usepackage{amsmath,amssymb,amsfonts,verbatim}
\usepackage{algorithmic}
\usepackage{graphicx,color,subcaption}
\usepackage{textcomp}
\usepackage{xcolor}
\def\BibTeX{{\rm B\kern-.05em{\sc i\kern-.025em b}\kern-.08em
    T\kern-.1667em\lower.7ex\hbox{E}\kern-.125emX}}
\usepackage{	latexsym,%
		    	amsmath,%
			amssymb,%
			amsthm,
			graphicx,
			relsize,
			caption,
			accents,
			algorithmic,
			caption,
			subcaption,
			tikz,
			pgfplots,
			bm,
			amsfonts,
			mathtools,
			textcomp,
   enumerate
			}

\usepackage[section]{placeins}
\usepackage[inline]{enumitem}
\usepackage[utf8]{inputenc}
\usepackage{array}
\usepackage{cite}
\usepackage[ruled,vlined]{algorithm2e}

\pgfplotsset{compat=1.5.1}
\usetikzlibrary{patterns}
\usetikzlibrary{shapes.geometric}
\usetikzlibrary{shapes,snakes,patterns}
\usetikzlibrary{arrows,positioning,automata,calc}
\usetikzlibrary{plotmarks}

\definecolor{darkblue}{rgb}{0.0, 0.0, 0.55}
\usepackage{microtype}
\usepackage{graphicx}
\usepackage{booktabs} 

\usepackage[colorlinks=true]{hyperref}
\hypersetup{
linkcolor = darkblue,
citecolor = green!50!black,
}

\DeclareMathOperator*{\argmin}{arg\,min}

\newcolumntype{L}[1]{>{\raggedright\let\newline\\\arraybackslash\hspace{0pt}}m{#1}}
\newcolumntype{C}[1]{>{\centering\let\newline\\\arraybackslash\hspace{0pt}}m{#1}}
\newcolumntype{R}[1]{>{\raggedleft\let\newline\\\arraybackslash\hspace{0pt}}m{#1}}

\newtheorem{theorem}{Theorem}

\allowdisplaybreaks[2]

\newtheorem{proposition}{Proposition}

\newtheorem{assumption}{Assumption}
\newtheorem{remark}{Remark}

\newcommand*{\rv}[1]{\mathsf{#1}}
\newcommand*{\vvec}[1]{\mathbf{#1}}
\newcommand*{\sset}[1]{\mathcal{#1}}
\newcommand*{\Rv}[1]{\boldsymbol{\mathsf{#1}}}

\newif\ifcomment
\commenttrue

\newif\ifcommentLater
\commentLaterfalse

\allowdisplaybreaks[2]

\begin{document}

\title{Approximate Gradient Coding for Privacy-Flexible Federated Learning with Non-IID Data}

\author{
\IEEEauthorblockN{Okko Makkonen\IEEEauthorrefmark{1}, Sampo Niemel\"{a}\IEEEauthorrefmark{1}, Camilla Hollanti\IEEEauthorrefmark{1}, Serge Kas Hanna\IEEEauthorrefmark{2}}
\IEEEauthorblockA{\IEEEauthorrefmark{1}%
Department of Mathematics and Systems Analysis, School of Science, Aalto University, Finland 
\IEEEauthorblockA{\IEEEauthorrefmark{2}%
I3S laboratory, C\^{o}te d'Azur University and CNRS, Sophia Antipolis, France \\
Emails: \{okko.makkonen, sampo.niemela, camilla.hollanti\}@aalto.fi, serge.kas-hanna@\{univ-cotedazur.fr, cnrs.fr\}
} } }

\maketitle


\begin{abstract}
This work focuses on the challenges of non-IID data and stragglers/dropouts in federated learning. We introduce and explore a privacy-flexible paradigm that models parts of the clients’ local data as non-private, offering a more versatile and business-oriented perspective on privacy. Within this framework, we propose a data-driven strategy for mitigating the effects of label heterogeneity and client straggling on federated learning. Our solution combines both offline data sharing and approximate gradient coding techniques. Through numerical simulations using the MNIST dataset, we demonstrate that our approach enables achieving a deliberate trade-off between privacy and utility, leading to improved model convergence and accuracy while using an adaptable portion of non-private data.
\end{abstract}


\section{Introduction}
In the ever-evolving landscape of machine learning (ML), one paradigm has emerged as a promising solution to balance the demand for data privacy and the need for robust distributed model training: federated learning (FL)~\cite{mcmahan2017communication, kairouz2021advancesSHORT}. This innovative approach allows multiple devices or entities, often called \emph{clients}, to collaboratively train a shared model under the orchestration of a central server while keeping their data decentralized. The result is a model that benefits from the collective intelligence of diverse data sources without compromising individual data privacy. Traditional distributed learning~\cite{jakovetic2013distributed,li2014scaling}, on the other hand, involves a central server that owns all the data and has complete control over how it distributes it to edge devices to parallelize the learning process, making it more suitable for scenarios where the data is naturally centralized.

While various assumptions can be made when designing algorithms for distributed learning and analyzing their performance, the most prominent split is between assuming independent, identically distributed (IID) data and non-IID data. The IID data assumption refers to the case where the data samples across different clients are independently drawn from the same underlying distribution, i.e., share similar statistical properties. This assumption simplifies the training process and algorithm design and enables deriving theoretical convergence guarantees such as the ones in~\cite{zhou2017convergence,stich2018local,wang2021cooperative,woodworth2018graph}. While this assumption may be valid in traditional distributed learning settings where the data is centralized, it almost never holds in practical FL settings since the clients independently collect their own training data, which typically vary in both size and distribution.

In the context of FL, the term non-IID data refers to a scenario where the data available across clients is not statistically similar. Previous studies have shown that dealing with non-IID data is one of the most significant hurdles encountered by FL. More specifically, training on non-IID data introduces several drawbacks that can hinder the effectiveness of collaborative model training, such as slower convergence and lower model accuracy~\cite{hsu2019measuring,non-IID,karimireddy2020scaffold,shi2022towards}.  The existing solutions in the literature for dealing with non-IID data in FL can be divided into two main categories that propose orthogonal approaches: {\em algorithm-based} methods and {\em data-driven} methods. Algorithm-based methods focus on adjusting the local loss function to align the local model with the global one\cite{wang2020tackling,wang2020federated,karimireddy2020scaffold,li2021model,durmus2021federated,shi2022towards}, training of personalized models for individual participants instead of employing a uniform global model \cite{smith2017federated,li2021ditto,fallah2020personalized,deng2020adaptive,liang2020think,mansour2020three}, and developing new aggregation schemes to enhance the model aggregation process\cite{hsu2019measuring,wang2020federated,wang2020tackling,yurochkin2019bayesian}. Data-driven approaches predominantly focus on data augmentation methods that mitigate statistical imbalances by artificially expanding the local dataset of each client through generating synthetic data~\cite{zhang2017mixup,jeong2018communication,goetz2020federated,rasouli2020fedgan,hao2021towards,yoon2021fedmix}. Other data-driven approaches include {\em  hybrid} federated learning schemes where a statistically diverse share of the training data is presumed to be present at the central server. In these hybrid schemes, the central server is expected to actively engage in the training process to compensate for potential statistical imbalances arising from client-side distributed computations\cite{non-IID,yoshida2020hybrid}.

In addition to the statistical challenge of dealing with non-IID data, FL also suffers from the straggler problem. The straggler or dropout problem refers to the issue where some participating clients or devices temporarily drop out or fail to complete their participation in a FL round. This can happen because clients may join or leave the FL system dynamically or due to other practical considerations such as poor network connections and resource constraints. Previous studies have demonstrated that the presence of stragglers aggravates the problem of learning on non-IID data, resulting in diminished model quality and slower convergence~\cite{charles2020outsized,mitra2021linear}. Coding for straggler mitigation is a popular solution that has been extensively studied in the traditional distributed learning setting~\cite{TLDK17,YA18,lee2018speeding,karakus2017straggler,Ozfatura2020,charles2017approximate,bitar2020stochastic,wang2019erasurehead,wang2019fundamental,glasgow2021approximate}, and also more recently in FL~\cite{prakash2020coded,schlegel2023codedpaddedfl}. The key idea in these works is to first introduce training data redundancy in the distributed system and then apply coding techniques that exploit this redundancy to either approximate or recover the central model update, even when some clients drop out. 

While previous works in FL treat all the clients' local data as equally private, our work introduces and investigates a {\em privacy-flexible} FL paradigm that allows for a deliberate trade-off between privacy and utility, offering more adaptable privacy measures. Our motivation for studying this setting stems from practical and commercial situations where striking the right balance between privacy and utility is essential for integrating FL in business models. More precisely, in the proposed paradigm, we model a portion of each client's data as being non-private.  We argue that parts of the local data in FL can be treated as non-private for many reasons, including: \begin{enumerate*}[label={\em (\roman*)}]
\item Participants often have diverse datasets of varying sensitivity, and some data may not contain anything private or sensitive, allowing for differentiation in privacy treatment. 
\item Privacy can be {\em selective}, i.e., some participants may be incentivized to sacrifice their privacy by revealing or sharing part of their raw data for enhancing the model's performance and/or for potential financial rewards.   \end{enumerate*} 

In this work, we introduce a novel data-driven approach that leverages the concept of privacy flexibility to improve the utility of FL while addressing the challenges posed by non-IID data and stragglers. As data can be non-IID in different ways, we specifically address {\em label heterogeneity}, which refers to the unequal representation of certain classes or labels across clients. Our proposed scheme consists of two key components. First, we introduce an offline {\em data sharing} mechanism, executed just once before the training phase. In this process, participants share some of their non-private data with each other to reduce the statistical imbalances resulting from label heterogeneity. The data sharing also creates redundancy in the training datasets as some of the data becomes available to multiple clients. The second component of our scheme capitalizes on this redundancy via {\em approximate gradient coding}. This coding method is designed to optimize the training process by providing an unbiased estimate of the central model update rule of gradient descent in the presence of stragglers, similar to the approach in~\cite{bitar2020stochastic} for traditional distributed learning settings. Moreover, we theoretically demonstrate that our proposed scheme reduces the variance of the obtained estimate, suggesting faster convergence, and characterize this reduction in terms of system parameters. Additionally, we present simulation results on real data using the MNIST dataset~\cite{deng2012mnist}, which substantiate our theoretical findings. These simulations reveal that our scheme enables achieving a convergence behavior and model accuracy that closely mirrors the IID case, contingent on the amount of shared non-private data.

\section{Preliminaries}
\label{sec:system_model}

\subsection{Notation} We use bold letters for vectors and sans-serif letters for random variables, e.g., $\boldsymbol{x}$, $\rv{X}$, and $\Rv{X}$ represent a vector, a random variable, and a random vector, respectively. Uppercase italic letters are used for sets. Let $[n]\triangleq \{1,2,\ldots,n\}$ be the set of integers from $1$ to $n$ (inclusive), and let \mbox{$\Delta^{n}\triangleq \big\{(\delta_0,\delta_1,\ldots,\delta_n)\in \mathbb{R}^{n+1} \big| \sum_{i=0}^{n} \delta_i=1~\text{and}~\delta_i\geq 0\big\}$} be the standard $n$-simplex. We represent the Euclidean norm of a vector $\boldsymbol{x}$ by $\|\boldsymbol{x}\|$, and the scalar product between two vectors $\boldsymbol{x}$ and $\boldsymbol{y}$ by $\langle\boldsymbol{x}, \boldsymbol{y} \rangle$.

\subsection{System Model} Consider a FL setting with $N>1$ clients who want to use their local data to collectively train a machine learning model with the help of a central server. Suppose that the clients want to run a supervised classification task, i.e., each client has a set of training examples $(\boldsymbol{x}, y)$ with features $\boldsymbol{x}$ and labels $y \in [L]$, where $L$ represents the total number of classes. Let $\mathcal{D}_i$ be the local dataset of client $i \in [N]$, and let \mbox{$\mathcal{D} = \bigcup_{i=1}^N \mathcal{D}_i$} denote the global training dataset of size $M$ given by the union of the local datasets across all the clients. The goal of FL is to find an optimal global model $\boldsymbol{\beta}^*$ that minimizes a given loss function~$\sset{L}$, i.e.,
\begin{equation*}
    \boldsymbol{\beta}^* = \argmin_{\boldsymbol{\beta}} \sset{L}(\mathcal{D}, \boldsymbol{\beta}).
\end{equation*}
The global loss can be often expressed as the sum of the individual losses evaluated at each example $(\boldsymbol{x}, y)$, i.e.,
\begin{equation*}
    \sset{L}(\mathcal{D},\boldsymbol{\beta}) = \sum_{(\boldsymbol{x}, y) \in \mathcal{D}} \sset{L}(\boldsymbol{x}, y, \boldsymbol{\beta}).
\end{equation*}
A common approach for solving such optimization problems is using gradient-based methods. Namely, the central server applies or approximates the following gradient descent update rule at each iteration $t \in \{0, 1, \ldots\}$ of the algorithm
\begin{equation*}
    \boldsymbol{\beta}^{(t+1)} = \boldsymbol{\beta}^{(t)} - \frac{\eta}{M} \sum_{(\boldsymbol{x}, y) \in \mathcal{D}}\nabla \sset{L}(\boldsymbol{x}, y, \boldsymbol{\beta}^{(t)}),
\end{equation*}
where $\nabla \sset{L}$ is the gradient of the loss function and $\eta$ represents the learning rate. In distributed gradient descent, the central server first sends the current model $\boldsymbol{\beta}^{(t)}$ to all clients. Each client $i$ then locally computes the individual gradients of the loss function over the examples in its local dataset $\mathcal{D}_i$ and sends a linear combination of these gradients to the central server. The central server then aggregates these local computations and either recovers the full gradient $\nabla\sset{L}(\mathcal{D},\boldsymbol{\beta}^{(t)})$ or obtains an estimate of it. The previous steps are repeated iteratively until convergence. If some of the clients are unresponsive (straggler/dropout) during certain iterations, the central server will not be able to recover the full gradient. In this case, the master's aggregate gives an estimate of the full gradient. In this work, we consider a straggling model where each client is unresponsive with probability $p$ in any given iteration, where this probability is independent across clients and iterations.

Suppose that the global training dataset $\mathcal{D}$ consists of a total of $M$ training examples such that there are $K_\ell$ examples corresponding to class $\ell \in [L]$. Let us assume that $K_\ell = K$ for all $\ell \in [L]$, such that $M = KL$. Let $\rv{X}_i^\ell$ denote the proportion of instances of label $\ell \in [L]$ that are owned by client $i \in [N]$. We focus on a label-heterogeneous setting where the initial proportions $\Rv{X}^\ell = (\rv{X}_1^\ell, \ldots, \rv{X}_N^\ell) \in \Delta^{N - 1}$ are typically far from \mbox{$\mathbf{\Theta}^{\star} \triangleq \left(\frac{1}{N}, \frac{1}{N}, \ldots, \frac{1}{N} \right) \in \Delta^{N-1}$}, where $\mathbf{\Theta}^{\star}$ corresponds to the perfectly label-homogeneous setting. We adopt the squared Euclidean distances between the label proportions and $\mathbf{\Theta}^{\star}$ as a measure for label heterogeneity. Some examples for modeling label-heterogeneity are the following:
\begin{enumerate}[leftmargin=*]
\item The single-class label-heterogeneous setting, where each client has data belonging only to a single class~\cite{non-IID}. For instance, for every $\ell \in [L]$, the initial label proportions could be \mbox{$\Rv{X}^\ell = (\rv{X}_1^\ell, \rv{X}_2^\ell, \ldots, \rv{X}_N^\ell) = \mathbf{e}^\ell$}, where $\mathbf{e}^\ell$ is the $\ell^{th}$ standard basis vector of~$\mathbb{R}^N$, with $N = L$. In this case, $\Rv{X}^\ell$ is deterministic, with $\|\Rv{X}^\ell - \mathbf{\Theta}^{\star} \|^2=\frac{N-1}{N}$ for all $\ell\in[L]$.
\item The initial label distributions are random and follow a Dirichlet distribution with a small concentration parameter~\cite{yurochkin2019bayesian}. Namely, for every $\ell \in [L]$, \mbox{$\Rv{X}^\ell = (\rv{X}_1^\ell, \rv{X}_2^\ell, \ldots, \rv{X}_N^\ell) \sim \operatorname{Dir}_N(\alpha)$}, where $\operatorname{Dir}_N(\alpha)$ is the Dirichlet distribution with $N$ categories and $\alpha > 0$ is the concentration parameter. Smaller values of $\alpha$ (close to zero) correspond to strongly heterogeneous settings; while $\alpha = \infty$ corresponds to a perfectly homogeneous one.
\end{enumerate}

Furthermore, we propose and study a privacy-flexible setting where a proportion $c \in [0,1]$ of each client's data is considered to be non-private. This proportion is assumed to be evenly distributed across all classes, i.e., each client $i \in [N]$ has $\rv{C}_i^\ell \triangleq c \rv{X}_i^\ell K$ non-private training examples from class $\ell \in [L]$.\footnote{To simplify notation, we assume that $c\rv{X}_i^\ell K$ are integers for all $i$ and $\ell$.} We assume that clients collect their local data independently, which implies that disclosing the non-private data of one client does not leak information about the private data of another client. 

\section{Proposed Scheme} \label{scheme}

To mitigate the effects of label heterogeneity and client straggling, we propose using an offline data sharing scheme that generates redundancy across clients by replicating non-private data. We will focus on the proportions of some fixed label \mbox{$\ell\in[L]$} and drop the superscript $\ell$. Recall that $\Rv{X} = (\rv{X}_1, \ldots, \rv{X}_N)$ denotes the initial label proportions prior to any data sharing, and let $\Rv{Y} = (\rv{Y}_1, \ldots, \rv{Y}_N)$ denote the corresponding final label proportions after data sharing. The replication-based data sharing scheme is described by $\Rv{S}=(\rv{S}_1,\ldots,\rv{S}_N)\in \mathbb{Z}^N$, where $\rv{S}_i\geq 0$ denotes the number  of non-private training examples that client~$i \in [N]$ receives from other clients.\footnote{Similar to~\cite{schlegel2023codedpaddedfl}, client-to-client offline communication can be assumed to be routed through the central server to ensure network connectivity.} Then, 
\begin{equation} \label{eq1}
    \Rv{Y} = \frac{\Rv{X}K + \Rv{S}}{K + B},
\end{equation}
where $B=\sum_{i=1}^{N} \rv{S}_i$ denotes the total amount of shared data.

The goal of data sharing is twofold:
\begin{enumerate*}[label={\em (\roman*)}]
\item ``Break" label heterogeneity by generating final label proportions $\Rv{Y}$ that are as close as possible to $\mathbf{\Theta}^{\star}$ in order to reduce the effects of ``non-IIDness". \item Create data redundancy across clients which we will use to ensure resilience against straggling clients.
\end{enumerate*}

As previously mentioned, we use the squared Euclidean distances between the label proportions and $\mathbf{\Theta}^{\star}$ as a measure for label heterogeneity. Prior to any data sharing the squared distance is $\lVert \Rv{X} - \mathbf{\Theta}^{\star} \rVert^2$, and after data sharing, the squared distance becomes $\lVert \Rv{Y}-\mathbf{\Theta}^{\star} \rVert^2$, where $\Rv{Y}$ is given by~\eqref{eq1}. We aim to devise a data sharing scheme, i.e., determine $\Rv{S}$, that minimizes $\rVert \Rv{Y} - \mathbf{\Theta}^{\star} \rVert^2$ for a given realization $\boldsymbol{X}$ of $\Rv{X}$. If the label counts of each client's local data $\mathcal{D}_i$ (private and non-private data) are assumed to be public, one can obtain an optimal deterministic data sharing scheme that minimizes $\rVert \Rv{Y} - \mathbf{\Theta}^{\star} \rVert^2$ for any realization $\boldsymbol{X}$ of $\Rv{X}$ by solving a constrained optimization problem. However, knowing the label counts of each client's data can potentially leak some information about the private data of the clients. Therefore, we propose a randomized data sharing scheme in Section~\ref{randshare} that does not require the knowledge of the label counts. We provide theoretical guarantees on the performance of the randomized data sharing scheme by evaluating $\mathbb{E}\left[\|\Rv{Y}-\mathbf{\Theta}^{\star} \|^2 \mid \Rv{X}=\boldsymbol{X}\right]$ in terms of the initial squared distance $\|\boldsymbol{X}-\mathbf{\Theta}^{\star}\|^2$, where the expectation is taken over the randomness of the data sharing process. Furthermore, in addition to minimizing the effects of label heterogeneity, an equally important consequence of the proposed data sharing scheme is generating redundancy across clients. In Section~\ref{app}, we present an approximate gradient coding scheme that leverages this redundancy to provide resilience against straggling clients.

\subsection{Randomized Data Sharing} \label{randshare}

To reduce the effects of label heterogeneity without the knowledge of the label counts of each client's local data, we propose using a randomized offline data sharing scheme where each client shares its non-private data with \mbox{$d \in \{0, 1, \ldots, N-1\}$} other clients chosen uniformly at random. Let \mbox{$\widetilde{\rv{C}}_i=\sum_{i'\neq i}\rv{C}_{i'}=c(1-\rv{X}_i)K$} denote the total number of non-private examples in $\mathcal{D}$  that are not owned by client $i\in [N]$. The randomized data sharing scheme places each of the $\widetilde{\rv{C}}_i$ non-private examples independently with probability $\frac{d}{N-1}$ at client $i \in [N]$. Thus, client $i\in[N]$ receives $\rv{S}_i$ new examples, where \mbox{$\rv{S}_i \sim \operatorname{Binomial}(\widetilde{\rv{C}}_i, \frac{d}{N-1})$} is a random variable that follows a binomial distribution with parameters $\widetilde{\rv{C}}_i=c(1-\rv{X}_i)K$ and $\frac{d}{N-1}$. Hence, the new label proportions after the randomized data sharing follow from~\eqref{eq1}, with $B = dcK$.

The next theorem expresses the expected value of $\|\Rv{Y}-\mathbf{\Theta}^{\star} \|^2$ in terms of the replication factor $d \in \{0, 1, \ldots, N-1\}$, privacy parameter $c \in [0, 1]$, number of clients $N$, and the initial distance $\|\boldsymbol{X}-\mathbf{\Theta}^{\star} \|^2$ for any realization $\boldsymbol{X}\in \Delta^{N-1}$ of $\Rv{X}$. The proof of this theorem is given in Appendix~\ref{app1}.

\begin{theorem} \label{thm1}
For any realization $\boldsymbol{X}\in \Delta^{N-1}$ of $\Rv{X}$, the randomized data sharing generates label proportions satisfying
\begin{equation*}
    \mathbb{E}\left[\|\Rv{Y}-\mathbf{\Theta}^{\star} \|^2 \mid \Rv{X}=\boldsymbol{X}\right] = \frac{dc(N-1-d)}{(1+dc)^2(N-1)K} + \frac{(N-1-dc)^2}{(1+dc)^2(N-1)^2} \|\boldsymbol{X}-\mathbf{\Theta}^{\star} \|^2,
\end{equation*}
where the expectation is over the randomness of the data sharing scheme. Furthermore, for $K\gg 1$, we have
\begin{equation*}
    \mathbb{E}\left[\|\Rv{Y}-\mathbf{\Theta}^{\star} \|^2 \mid \Rv{X}=\boldsymbol{X}\right]\approx \frac{(N-1-dc)^2}{(1+dc)^2(N-1)^2} \|\boldsymbol{X}-\mathbf{\Theta}^{\star} \|^2.
\end{equation*}
\end{theorem}

The significance of this theorem is that for any given set of input label proportions $\boldsymbol{X}$, the expected value of the output distance $\|\Rv{Y}-\mathbf{\Theta}^{\star} \|^2$ is reduced by at least a factor of $(1+dc)^2$ with respect to the initial distance $\|\boldsymbol{X}-\mathbf{\Theta}^{\star} \|^2$ for large enough~$L$. This shows that the randomized data sharing scheme can effectively reduce heterogeneity without knowing the label counts of each client's local data.

\subsection{Approximate Gradient Coding} \label{app}
As previously mentioned, at each iteration $t \in \{0, 1, \ldots\}$, the central server sends the current model $\boldsymbol{\beta}^{(t)}$ to all $N$ clients. Each client participates in any given iteration independently with probability $1 - p$, i.e., the client is a straggler with probability $p$. Let $\sset{S}_i\subseteq [M]$, $i \in [N]$, be the set of indices of the training examples $(\boldsymbol{x}, y) \in \sset{D}_i$ that are owned by client~$i$. The clients participating in a given iteration $t$ send a linear combination of the partial gradients computed over their local data which is given by
\begin{equation}
    f_i(\boldsymbol{\beta}^{(t)}) = \sum_{j \in \sset{S}_i} W_{j} \vvec{g}^{(t)}_j,
\end{equation}
where $\vvec{g}^{(t)}_j \triangleq \nabla \sset{L}(\boldsymbol{x}_j, y_j, \boldsymbol{\beta}^{(t)})$ denotes the partial gradient evaluated at example $(\boldsymbol{x}_j, y_j)$, and $W_{j}$ is the weighting factor of example~$(\boldsymbol{x}_j, y_j)$. The weighting factor of $(\boldsymbol{x}_j, y_j)$ is \mbox{$W_{j} = \frac{1}{(1-p)d_j}$}, where $d_j \in \mathbb{Z}^+$ is the total number of times the example $(\boldsymbol{x}_j, y_j)$ is replicated across all $N$ clients. Note that $d_j = 1$ for all private data, and $d_j=d+1$ for all non-private data, where $d$ is the replication factor of the randomized data sharing scheme. The central server then aggregates the local computations of the clients to obtain an estimate of the full gradient given by
\begin{equation} \label{est}
    \hat{\vvec{g}}^{(t)} = \sum_{i=1}^{N} \rv{I}_i^{(t)} f_i(\boldsymbol{\beta}^{(t)}),
\end{equation}
where $\rv{I}_i^{(t)} = 1$ if client $i$ is participating in iteration $t$, and $\rv{I}_i^{(t)} = 0$ otherwise. 
One can easily show that $\hat{\vvec{g}}^{(t)}$ is an unbiased estimator of the full gradient $\vvec{g}^{(t)} = \sum_{j=1}^{M} \vvec{g}^{(t)}_j$, i.e., \mbox{$\mathbb{E}[\hat{\vvec{g}}^{(t)}] = \vvec{g}^{(t)}$} (see Appendix~\ref{app2}). Note that the expectations considered in this section are over the randomness of the straggling process in iteration $t$, conditioned on the model $\boldsymbol{\beta}^{(t)}$ given by the central server to the clients. The variance of $\hat{\vvec{g}}^{(t)}$ depends on the data sharing scheme being used and has a direct effect on the rate of convergence of the algorithm. More specifically, it has been shown in~\cite{rakhlin2011making} that for an unbiased estimator, the value of $\mathbb{E}[\lVert \hat{\vvec{g}}^{(t)} \rVert^2]$ is inversely proportional to the rate of convergence under certain assumptions on the loss function. An important trait of our proposed scheme is that it reduces the variance of the estimator and thus speeds up the convergence of the algorithm. We theoretically demonstrate this phenomenon in Theorem~\ref{thm2} and illustrate it practically through numerical simulations in Section~\ref{simul}. For the single-class label-heterogeneous setting described in Section~\ref{sec:system_model}, Theorem~\ref{thm2} gives a lower bound on the expected difference between the variance of $\hat{\vvec{g}}^{(t)}$ before and after data sharing. The bound is expressed in terms of the replication factor $d \in \{0, 1, \ldots, N - 1\}$, privacy parameter $c \in [0,1]$, straggling probability $p\in[0,1)$, and other system parameters defined in Section~\ref{sec:system_model}. The proof of this theorem is given in Appendix~\ref{app2}.

\begin{theorem} \label{thm2}
Consider the single-class label-heterogeneous setting described in Section~\ref{sec:system_model} and let $\hat{\vvec{g}}^{(t)}_{\Rv{X}}$ be the estimate of the gradient obtained in iteration $t$ if no data is shared between clients. Let $\hat{\vvec{g}}^{(t)}_{\Rv{Y}}$ be the estimate of the gradient obtained in iteration~$t$ if the offline randomized data sharing scheme is applied with replication factor $d$ and privacy parameter $c$. Under Assumption~\ref{ass1} (Appendix~\ref{app2}), it holds that
\begin{equation*}
\mathbb{E}\left[\|\hat{\vvec{g}}^{(t)}_{\Rv{X}}\|^2-\|\hat{\vvec{g}}^{(t)}_{\Rv{Y}} \|^2\right] \geq \frac{p}{1-p} \frac{d-1}{d+1} \sum_{(j_1,j_2)\in \sset{J}_{\text{same}}^{\text{non-priv}}} \langle \vvec{g}^{(t)}_{j_1}, \vvec{g}^{(t)}_{j_2} \rangle,
\end{equation*}
where the expectation is over the randomness of the straggling process, and $\sset{J}_{\text{same}}^{\text{non-priv}}$ is the set of all index pairs $(j_1,j_2)\in [M]^2$ such that $(\boldsymbol{x}_{j_1},y_{j_1})$ and $(\boldsymbol{x}_{j_2},y_{j_2})$ are two examples belonging to the same class where at least one of them is non-private, with $\left \lvert \sset{J}_{\text{same}}^{\text{non-priv}}\right \rvert =MKc(2-c)$.
\end{theorem}

The result in Theorem~\ref{thm2} aligns with several intuitive aspects regarding the impact of the system parameters on the variance reduction achieved by the proposed scheme. For instance, the term $\frac{p}{1-p}$ indicates that the reduction in variance is more pronounced for higher values of $p$, which is intuitive since one would expect the benefits of data sharing to become more significant as the number of stragglers increases. Furthermore, the effects of the replication factor $d$ and privacy parameter $c$ are reflected by the term $\frac{d-1}{d+1}$ and the number of terms of the summation, respectively, where the latter is proportional to~$c(2-c)$. By analyzing these terms, one could show that for $d>1$ and a fixed offline communication cost characterized by the product~$dc$, higher values of $c$ lead to a greater reduction in variance. This observation is also intuitive, as relaxing the privacy constraint increases the diversity of redundancy generated by data sharing. 

Note that the theoretical analysis above relies on Assumption~\ref{ass1}, stated formally in Appendix~\ref{app2}. Informally, this assumption is based on the hypothesis that gradients computed on examples from the same class tend to align well in the feature space, resulting in large positive scalar products between these gradients. On the other hand, gradients from different classes may lack alignment, leading to smaller and potentially negative scalar products, reflecting the differences between the classes. This assumption is intuitive in practical scenarios where data in different classes are typically dissimilar, provided that training parameters are appropriately chosen to ensure proper convergence. A more detailed justification of this assumption is provided in Appendix~\ref{app2}, along with simulation results over real data supporting these claims.

\section{Simulation Results} \label{simul}

\subsection{Setup} \label{simul_setup}

We simulate a federated learning setup with $N=10$ clients. The goal is to train a multinomial logistic regression model on the MNIST dataset \cite{deng2012mnist}, which is a supervised image classification task consisting of $L=10$ different classes. We use a global training dataset $\mathcal{D}$ of size $M=300$, consisting of $K_{\ell}=K=30$ images from each class $\ell\in [L]$ drawn uniformly at random from the $60,000$ training images in MNIST. To model IID/non-IID settings, we consider multiple ways for partitioning the~$M=300$ training examples over the $N=10$ clients: \begin{enumerate*}[label={\em (\roman*)}]  \item IID setting, where the data is randomly shuffled, and then partitioned into $10$ clients each receiving $30$ examples. \item Two non-IID settings that follow from the single-class label-heterogeneous setting and the random Dirichlet distribution with $\alpha=0.1$, as described in Section~\ref{sec:system_model}. \end{enumerate*} 

We apply the randomized data sharing and approximate gradient coding to the two non-IID settings, and compare the performance of the trained model to the IID and non-IID cases with no data sharing. For each scenario, we run a total of 1000 simulations for 50 communication rounds (iterations) of the federated learning process and compute the average test accuracy and the second moment of the gradient estimator (defined in~\eqref{est}) as a function of the communication round. In each round, the identities of the stragglers/dropouts are determined according to the straggling model parameterized by $p$, as explained in Section~\ref{sec:system_model}. We train the model using the SGD (stochastic gradient descent) optimizer with a learning rate of $\eta = 0.1$ and decay~$\gamma = 0.97$. The source code for these simulations can be found in \cite{extended}.

\begin{figure*}[h!]
\centering
\begin{subfigure}[b]{.32\linewidth}
\includegraphics[width=\linewidth]{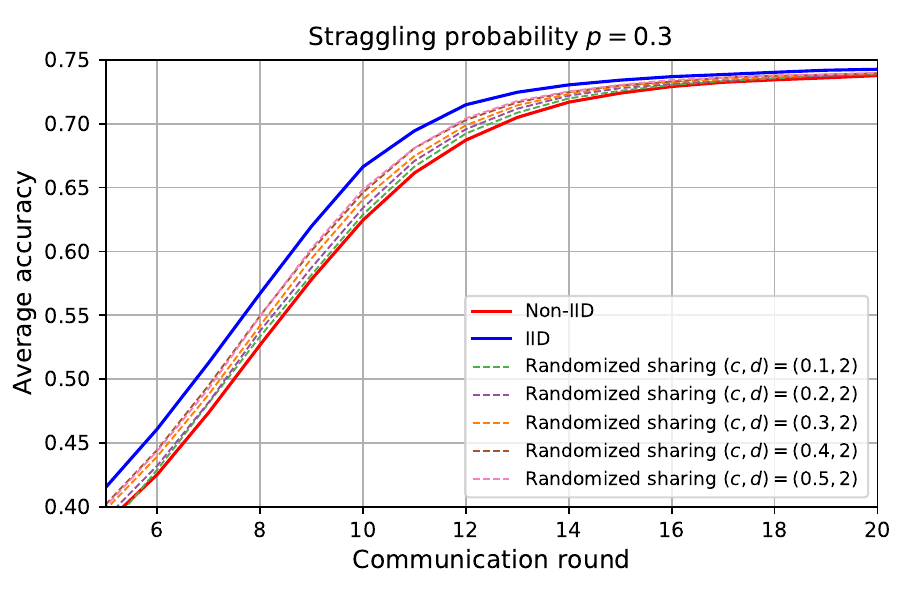}
\label{fig:dirichlet_accuracy_comparison_p_30}
\end{subfigure}
\begin{subfigure}[b]{.32\linewidth}
\includegraphics[width=\linewidth]{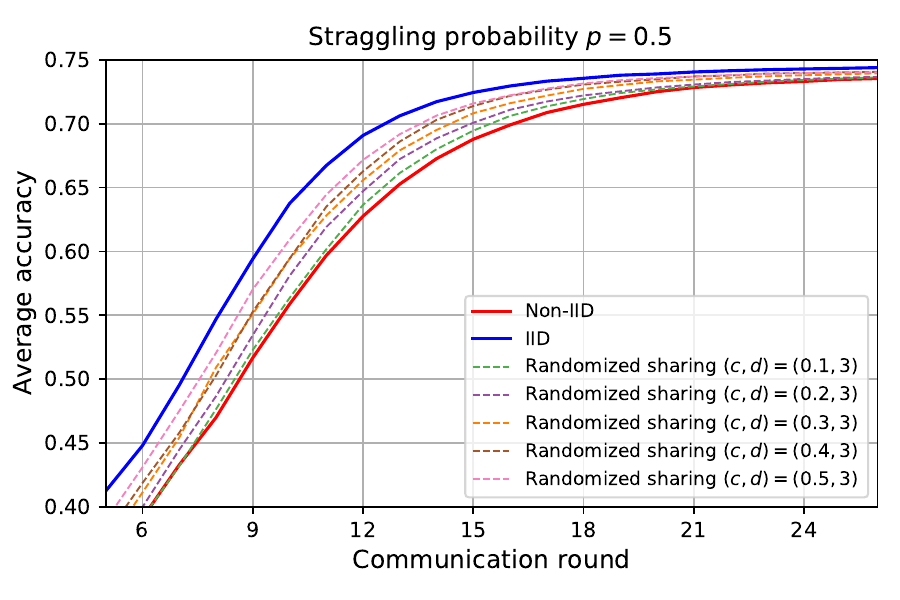}
\label{fig:dirichlet_accuracy_comparison_p_50}
\end{subfigure}
\begin{subfigure}[b]{.32\linewidth}
\includegraphics[width=\linewidth]{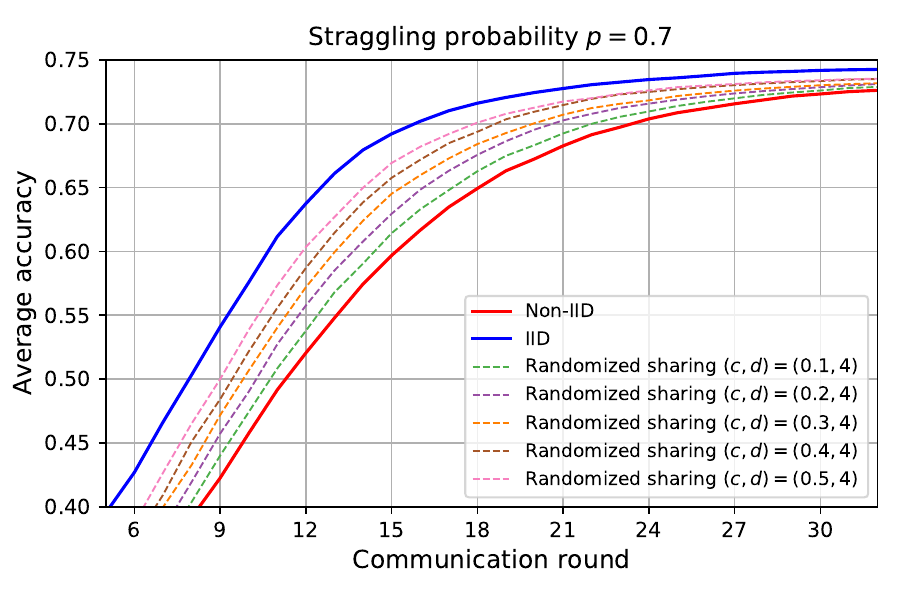}
\label{fig:dirichlet_accuracy_comparison_p_70}
\end{subfigure}

\begin{subfigure}[b]{.32\linewidth}
\vspace{-0.5cm}
\includegraphics[width=\linewidth]{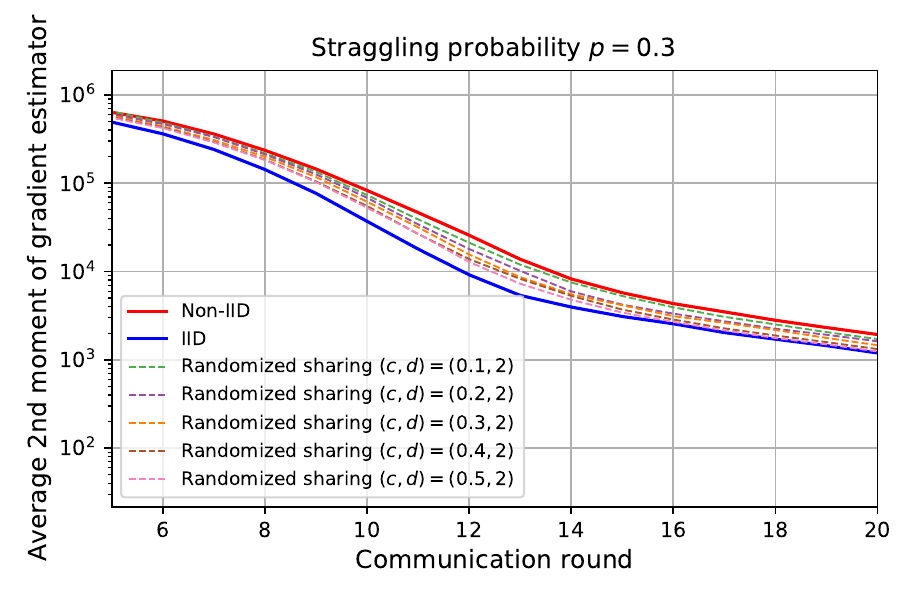}
\caption{Dirichlet setting $\alpha=0.1$, $p=0.3$}\label{fig:dirichlet_moments_comparison_p_30}
\end{subfigure}
\begin{subfigure}[b]{.32\linewidth}
\vspace{-0.5cm}
\includegraphics[width=\linewidth]{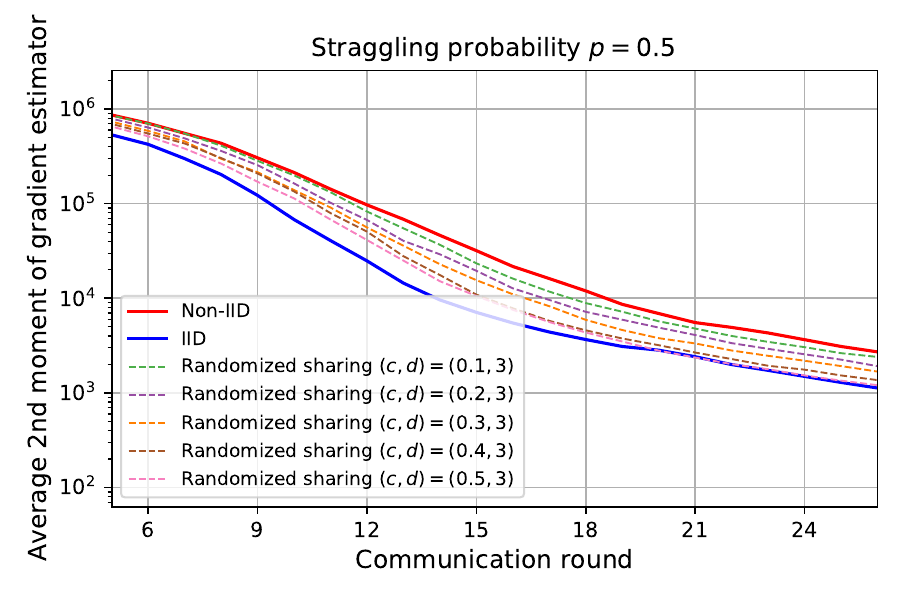}
\caption{Dirichlet setting $\alpha=0.1$, $p=0.5$}\label{fig:dirichlet_moments_comparison_p_50}
\end{subfigure}
\begin{subfigure}[b]{.32\linewidth}
\vspace{-0.5cm}
\includegraphics[width=\linewidth]{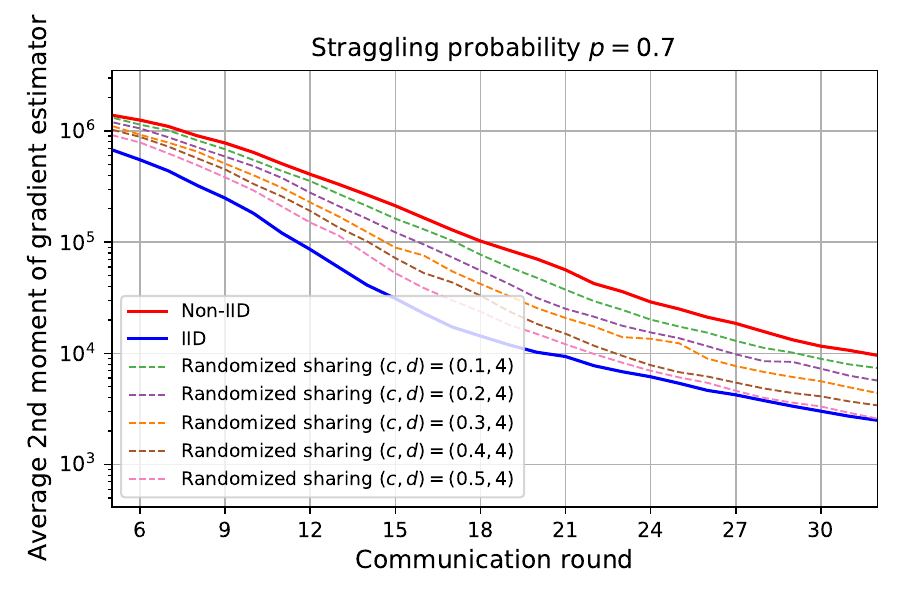}
\caption{Dirichlet setting $\alpha=0.1$, $p=0.7$}\label{fig:dirichlet_moments_comparison_p_70}
\end{subfigure}
\caption{Average testing accuracy and second moment of the gradient estimator (defined in~\eqref{est}) as a function of the communication round (iteration) in the Dirichlet setting with $\alpha = 0.1$.}
\label{fig:dirichlet_comparison}
\end{figure*}

\begin{figure*}[h!]
\centering
\begin{subfigure}[b]{.32\linewidth}
\includegraphics[width=\linewidth]{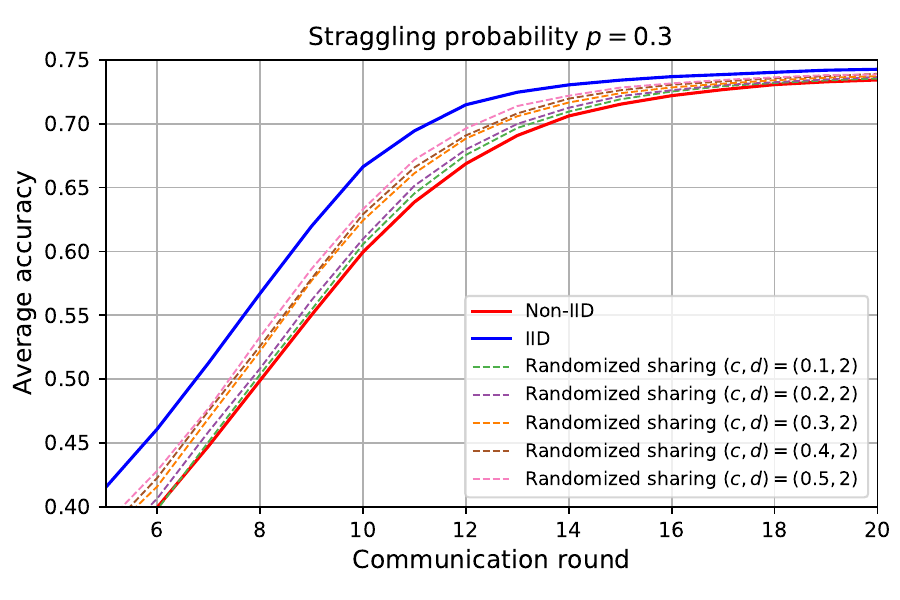}
\label{fig:heterogeneous_accuracy_comparison_p_30}
\end{subfigure}
\begin{subfigure}[b]{.32\linewidth}
\includegraphics[width=\linewidth]{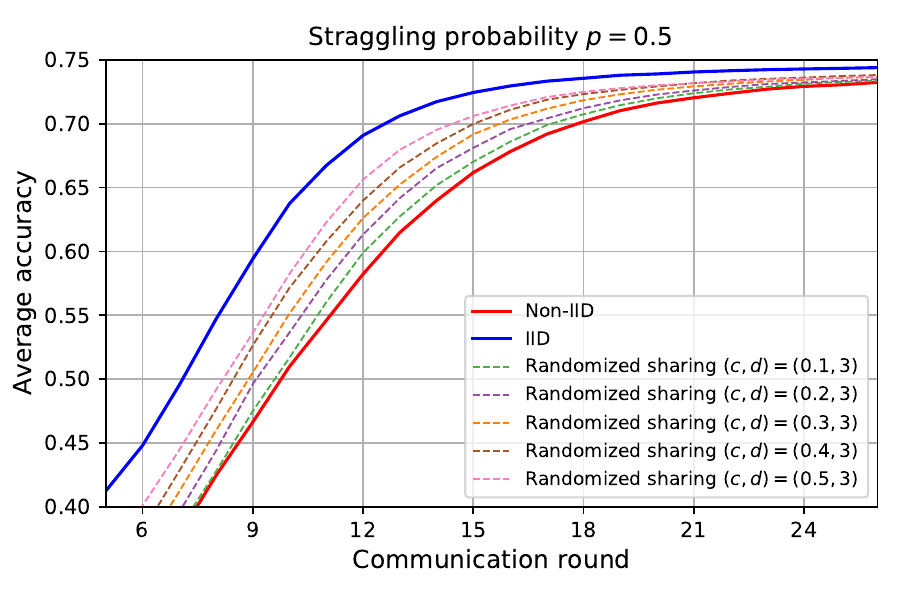}
\label{fig:heterogeneous_accuracy_comparison_p_50}
\end{subfigure}
\begin{subfigure}[b]{.32\linewidth}
\includegraphics[width=\linewidth]{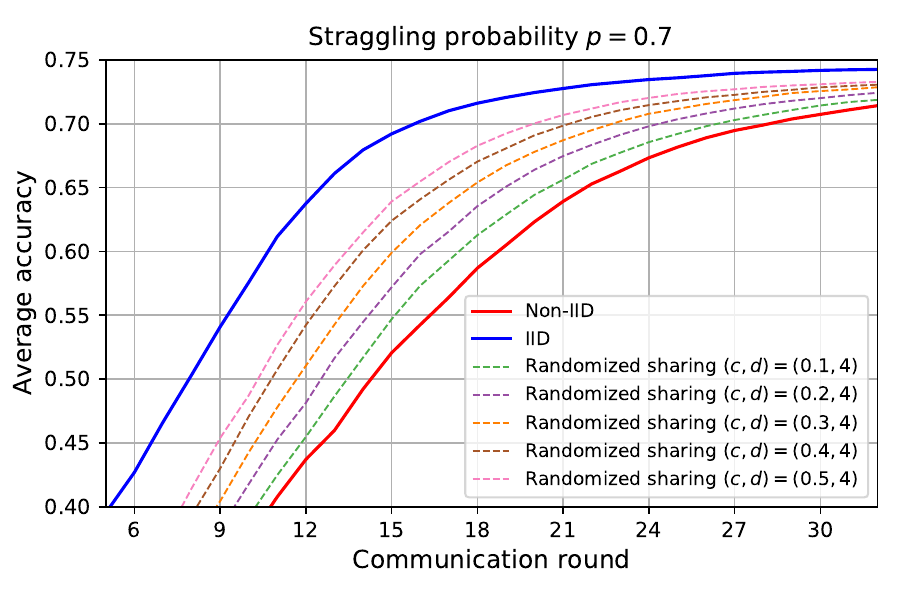}
\label{fig:heterogeneous_accuracy_comparison_p_70}
\end{subfigure}

\begin{subfigure}[b]{.32\linewidth}
\vspace{-0.5cm}
\includegraphics[width=\linewidth]{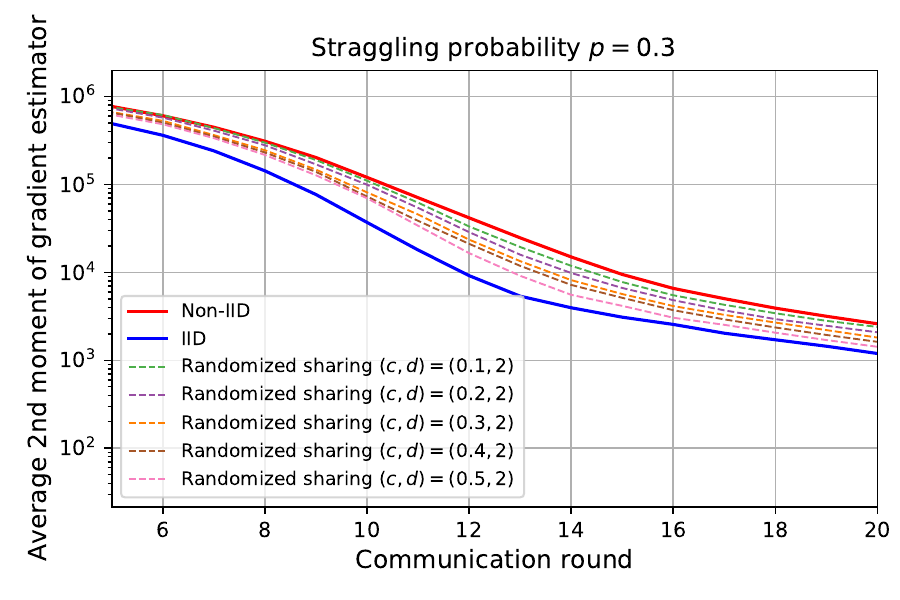}
\caption{Single-class setting, $p=0.3$}\label{fig:heterogeneous_moments_comparison_p_30}
\end{subfigure}
\begin{subfigure}[b]{.32\linewidth}
\vspace{-0.5cm}
\includegraphics[width=\linewidth]{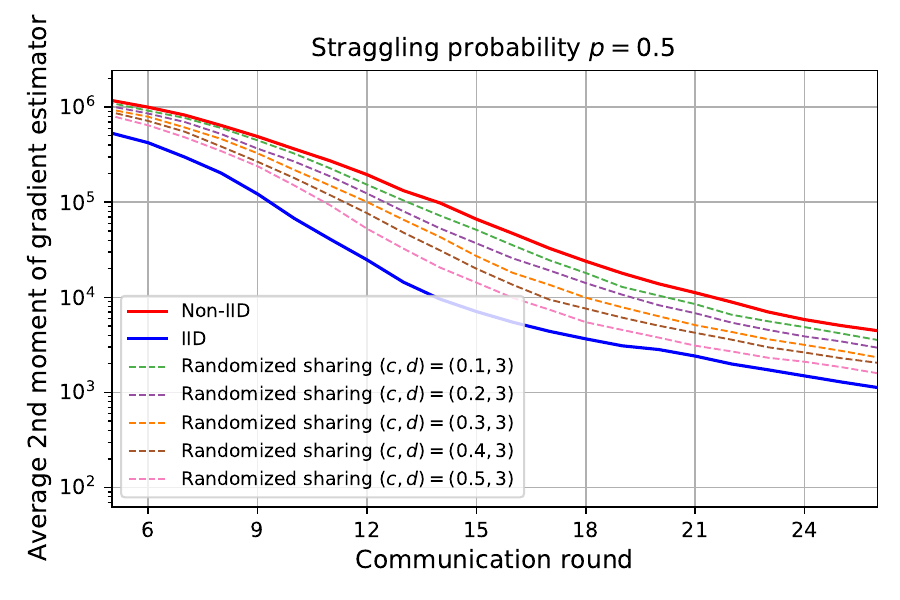}
\caption{Single-class setting, $p=0.5$}\label{fig:heterogeneous_moments_comparison_p_50}
\end{subfigure}
\begin{subfigure}[b]{.32\linewidth}
\vspace{-0.5cm}
\includegraphics[width=\linewidth]{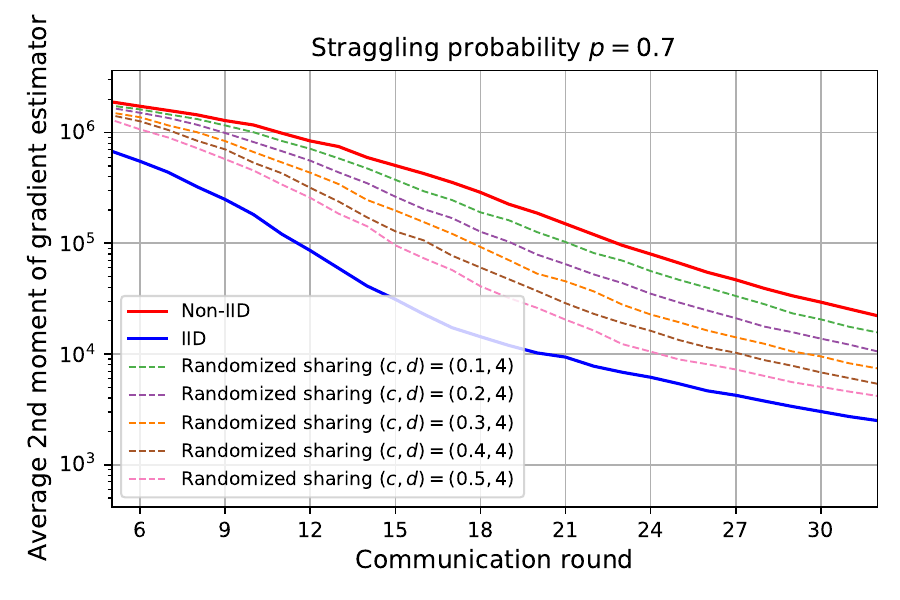}
\caption{Single-class setting, $p=0.7$}\label{fig:heterogeneous_moments_comparison_p_70}
\end{subfigure}
\caption{Average testing accuracy and second moment of the gradient estimator (defined in~\eqref{est}) as a function of the communication round (iteration) in the single-class setting.}
\label{fig:heterogeneous_comparison}
\end{figure*}

\subsection{Results}
The average testing accuracy over 1000 simulations is plotted in Figures~\ref{fig:dirichlet_comparison} and \ref{fig:heterogeneous_comparison} for different parameters and compared with the baseline non-IID (no data sharing) and IID settings. These simulations show that there is a clear increase in the convergence rate as the parameter $c$ is increased from $0$ to $0.5$. We also observe a slight difference in the final model accuracy achieved. We expect this difference to be more significant for more challenging datasets and more complex models. Using our proposed scheme and by manipulating the parameters $c$ and $d$, we may interpolate between the non-IID and the IID settings. Furthermore, the plots of the average second moment of the gradient estimator validate our intuition and theoretical analysis in Section~\ref{app}. Namely, the simulation results show that the increase in convergence rate is correlated with the decrease in the variance of the gradient estimator, which depends on the parameters of our scheme and the straggling behavior. By comparing Figures~\ref{fig:dirichlet_comparison} and \ref{fig:heterogeneous_comparison}, we see that the single-class heterogeneous setting leads to worse model convergence and accuracy than the Dirichlet setting with $\alpha = 0.1$. This suggests that given the single-class heterogeneous setting, one would need more data sharing to achieve the same performance as the Dirichlet case.

\section{Conclusion}
In conclusion, our theoretical and numerical results demonstrate that under the privacy-flexible FL paradigm that we introduce, combining data sharing with gradient coding enables a deliberate trade-off  between privacy (characterized by the parameter $c$) and utility (characterized by model convergence and accuracy). This trade-off is achieved at the expense of a one-time offline communication cost and additional local computation, which is a common price to pay when mitigating the effects of stragglers and non-IID data in distributed settings. In future work, we intend to analyze this trade-off for different models and datasets, and also other algorithms such as federated averaging, where multiple local model updates are performed in a single communication round.

\bibliographystyle{IEEEtran}
\bibliography{Refs}

\appendices

\section{Proof of Theorem 1} \label{app1} 
\noindent We have
\begin{align}
\mathbb{E}\left[\|\Rv{Y}-\mathbf{\Theta}^{\star} \|^2 \mid \Rv{X}=\boldsymbol{X}\right] &= \mathbb{E}\left[\sum_{i=1}^N \left(\rv{Y}_i-\frac{1}{N}\right)^2~\bigg|~\rv{X}_i=X_i\right] \\
&=\sum_{i=1}^N \mathbb{E}\left[ \rv{Y}_i^2| \rv{X_i}=X_i\right]-\frac{2}{N}\mathbb{E}\left[ \rv{Y}_i | \rv{X_i}=X_i\right] + \frac{1}{N^2}. \label{eq3}
\end{align}
Since $B=dcK$ for the randomized data sharing scheme, then it follows from~\eqref{eq1} that
\begin{equation}
    \mathbb{E}\left[ \rv{Y}_i \mid \rv{X_i}=X_i\right] = \frac{1}{1+dc} \left(X_i + \frac{\mathbb{E}[\rv{S}_i]}{K} \right) \\
    =  \frac{1}{1+dc} \left( X_i + \frac{dc}{N-1}(1-X_i) \right). \label{eq4}
\end{equation}
Furthermore,
\begin{align}
    \mathbb{E}\left[ \rv{Y}_i^2 \mid \rv{X_i}=X_i\right] &= \frac{1}{(1+dc)^2} \left(X_i^2 + 2\frac{\mathbb{E}[\rv{S}_i]}{K}X_i+ \frac{\mathbb{E}[\rv{S}_i^2]}{K^2}\right) \\
    &=  \frac{1}{(1+dc)^2} \left( X_i^2 + 2\frac{dc}{N-1}(1-X_i)X_i+ \frac{dc}{(N-1)K}\left(1-\frac{d}{N-1}\right)(1-X_i)+\frac{d^2c^2}{(N-1)^2}(1-X_i)^2 \right). \label{eq6}
\end{align}
By substituting~\eqref{eq4} and \eqref{eq6} in \eqref{eq3} we get
\begin{multline}
\mathbb{E}\left[\|\Rv{Y}-\mathbf{\Theta}^{\star} \|^2 \mid \Rv{X}=\boldsymbol{X}\right]= \frac{dc(N-1-d)}{(1+dc)^2(N-1)^2K}  \sum_{i=1}^N 1-X_i \\
+\frac{1}{(1+dc)^2} \underbrace{\left(1-2\frac{dc}{N-1}+\frac{d^2c^2}{(N-1)^2} \right)}_{T_1} \sum_{i=1}^N  X_i^2 \\
-\frac{2}{N} \frac{1}{(1+dc)^2} \underbrace{\left(1+dc+\frac{d^2c^2N}{(N-1)^2}-\frac{dcN}{N-1} - \frac{dc(1+dc)}{N-1} \right)}_{T_2} \sum_{i=1}^N  X_i \\
+ \frac{1}{N^2} \frac{1}{(1+dc)^2} \underbrace{\left((1+dc)^2-2\frac{(1+dc)dcN}{N-1}+\frac{d^2c^2N^2}{(N-1)^2} \right)}_{T_3}.
\end{multline}
One can easily verify that
\begin{equation}
    T_1=T_2=T_3=\frac{(N-1-dc)^2}{(N-1)^2}.
\end{equation}
Therefore,
\begin{align}
    \mathbb{E}\left[\|\Rv{Y}-\mathbf{\Theta}^{\star} \|^2 \mid \Rv{X}=\boldsymbol{X}\right]=\frac{dc(N-1-d)}{(1+dc)^2(N-1)^2K}  \sum_{i=1}^N 1-X_i + \frac{(N-1-dc)^2}{(1+dc)^2(N-1)^2} \sum_{i=1}^N \left(X_i-\frac{1}{N}\right)^2. \label{eq9}
\end{align}
Since $\boldsymbol{X}\in \Delta^{N-1}$, we have $\sum_{i=1}^N 1-X_i=N-1$. The proof is concluded by substituting the latter equality in the first term in~\eqref{eq9}, and by expressing the summation in the second term as $\|\boldsymbol{X}-\boldsymbol{\Theta}^*\|^2$.

\section{Proof of Theorem 2} \label{app2} 
\begin{figure}[h!] 
\centering
\begin{subfigure}[b]{.49\linewidth}
\includegraphics[width=\linewidth]{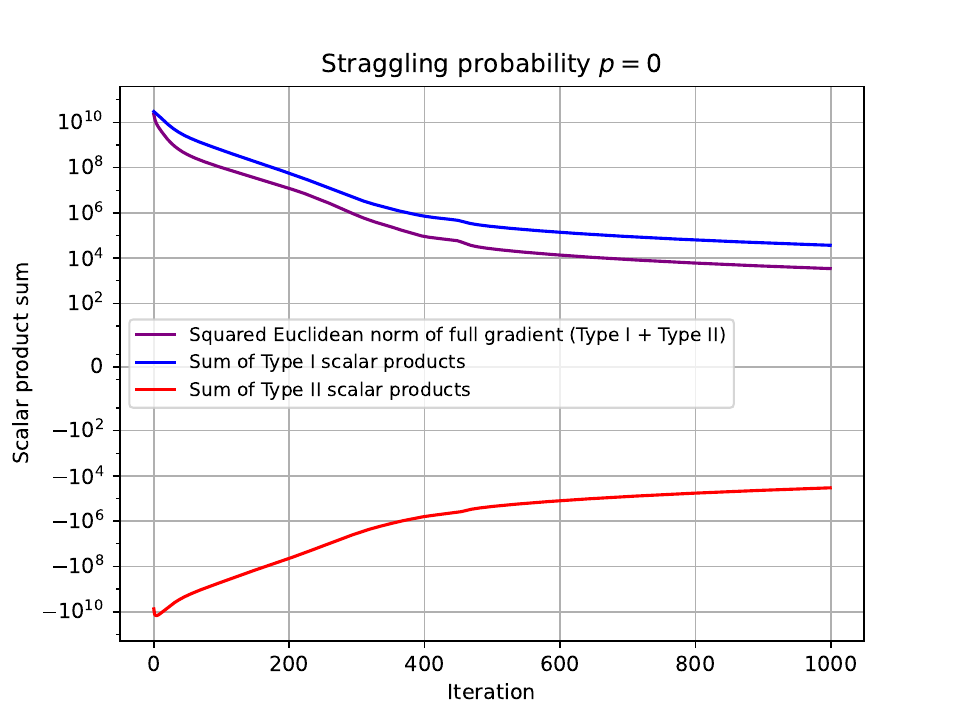}
\label{fig:scalar_comparison_p_0}
\vspace{-0.4cm}
\caption{No data sharing, $p=0$.}
\end{subfigure}
\begin{subfigure}[b]{.49\linewidth}
\includegraphics[width=\linewidth]{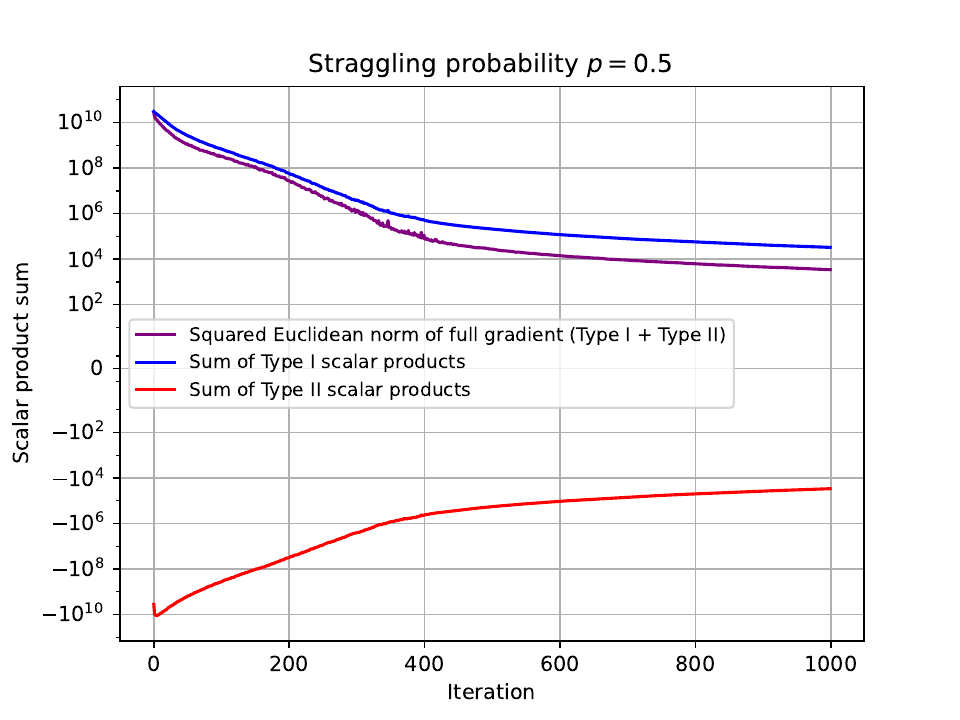}
\label{fig:scalar_comparison_50}
\vspace{-0.4cm}
\caption{Randomized data sharing, $(c,d)=(0.2,3)$ and $p=0.5$.}
\end{subfigure}
\caption{The same simulation setup, as explained in Section~\ref{simul_setup} based on the MNIST dataset, is considered under the single-class label-heterogeneous setting. The figure shows the sums of the scalar products of gradients computed on examples from the same and different classes, referred to as Type I and Type II scalar products, respectively. These sums are reported at each iteration $t$ of the algorithm. Additionally, the squared Euclidean norm of the full gradient $\mathbf{g}^{(t)}= \sum_{j=1}^{M} \vvec{g}^{(t)}_j$, obtained by summing both Type~I~and~II scalar products, is also given. A fixed learning rate of $\eta=0.01$ is used.}
\label{scalar_comparison}
\end{figure}

\noindent The gradient estimate obtained at iteration $t$ can be expressed as
\begin{equation*}
     \hat{\vvec{g}}^{(t)} =  \sum_{j=1}^M \frac{\rv{Z}_j^{(t)}}{(1-p)d_j} \vvec{g}^{(t)}_j,
\end{equation*}
where $d_j \in [N]$ is the number of replicates of example $(\boldsymbol{x}_j, y_j)$ that are available across the $N$ clients and $\rv{Z}_j^{(t)}\in [d_j]$ is the random variable representing the number of non-stragglers participating in iteration $t$ that have example $(\boldsymbol{x}_j, y_j)$. Suppose that $(\boldsymbol{x}_j, y_j)$ is replicated across $d_j$ different clients indexed by $i_1,i_2,\ldots,i_{d_j}$. Then, $\rv{Z}_j^{(t)}=\rv{I}_{i_1}^{(t)} + \rv{I}_{i_2}^{(t)} + \ldots + \rv{I}_{i_{d_j}}^{(t)}$ where $\rv{I}_i^{(t)} = 1$ if client $i$ is participating in iteration $t$, and $\rv{I}_i^{(t)} = 0$ otherwise. It follows from the straggling model that $\rv{I}_{i_1}^{(t)}, \rv{I}_{i_2}^{(t)}, \ldots ,\rv{I}_{i_{d_j}}^{(t)}$ are independent Bernoulli random variables, and hence $\rv{Z}_j^{(t)}$ is a binomial random variable with parameters $d_j$ and $1-p$. 
Therefore, by linearity of expectation, we have
\begin{equation} \label{unb}
    \mathbb{E} \left[ \hat{\vvec{g}}^{(t)} \right] = \sum_{j=1}^M \frac{\mathbb{E} \left[ \rv{Z}_j^{(t)}\right]}{(1-p)d_j} \vvec{g}^{(t)}_j = \sum_{j=1}^M \vvec{g}^{(t)}_j = \vvec{g}^{(t)},
\end{equation}
which proves that the estimator is unbiased. Recall that the expectation computed in~\eqref{unb} and henceforth is over the randomness of the straggling process in iteration $t$, conditioned on the model $\boldsymbol{\beta}^{(t)}$ given by the central server to the clients. Furthermore, the second moment of the gradient estimator is given by
\begin{equation} \label{eq:secm}
     \mathbb{E} \left[ \| \hat{\vvec{g}}^{(t)} \|^2 \right]=  \sum_{(j_1,j_2)\in [M]^2} \frac{\mathbb{E} \left[ \rv{Z}_{j_1}^{(t)}\rv{Z}_{j_2}^{(t)}\right]}{(1-p)^2d_{j_1}d_{j_2}} \langle \vvec{g}^{(t)}_{j_1}, \vvec{g}^{(t)}_{j_2} \rangle.
\end{equation}
The analysis of the second moment depends on how the $M$ training examples are distributed across the clients. Consider the single-class label-heterogeneous setting, and let  $\sset{J}_{\text{same}}$ and $\sset{J}_{\text{diff}}$ be the set of all index pairs $(j_1,j_2)\in [M]^2$ such that $(\boldsymbol{x}_{j_1},y_{j_1})$ and $(\boldsymbol{x}_{j_2},y_{j_2})$ are two examples belonging to the same class and different classes, respectively. Prior to data sharing, we have $d_{j_1}=d_{j_2}=1$ for all $(j_1,j_2)\in [M]^2$. Furthermore, if $(j_1,j_2)\in \sset{J}_{\text{same}}$, we have $\rv{Z}_{j_1}^{(t)}\rv{Z}_{j_2}^{(t)}=\rv{I}_{i}^{(t)}\rv{I}_{i}^{(t)}$ for some $i\in [N]$, else if $(j_1,j_2)\in \sset{J}_{\text{diff}}$, then $\rv{Z}_{j_1}^{(t)}\rv{Z}_{j_2}^{(t)}=\rv{I}_{i_1}^{(t)}\rv{I}_{i_2}^{(t)}$ for some $(i_1,i_2)\in [N]^2$ with $i_1\neq i_2$. Therefore,
\begin{align} 
     \mathbb{E} \left[ \| \hat{\vvec{g}}^{(t)}_{\Rv{X}} \|^2 \right] &=  \frac{\mathbb{E} \left[ \rv{I}_{i}^{(t)}\rv{I}_{i}^{(t)}\right]}{(1-p)^2} \sum_{(j_1,j_2)\in \sset{J}_{\text{same}}}  \langle \vvec{g}^{(t)}_{j_1}, \vvec{g}^{(t)}_{j_2} \rangle + \frac{\mathbb{E} \left[ \rv{I}_{i_1}^{(t)}\rv{I}_{i_2}^{(t)}\right]}{(1-p)^2} \sum_{(j_1,j_2)\in \sset{J}_{\text{diff}}}  \langle \vvec{g}^{(t)}_{j_1}, \vvec{g}^{(t)}_{j_2} \rangle, \\
     &= \frac{1}{1-p} \sum_{(j_1,j_2)\in \sset{J}_{\text{same}}}  \langle \vvec{g}^{(t)}_{j_1}, \vvec{g}^{(t)}_{j_2} \rangle + \sum_{(j_1,j_2)\in \sset{J}_{\text{diff}}}  \langle \vvec{g}^{(t)}_{j_1}, \vvec{g}^{(t)}_{j_2} \rangle. \label{secmX}
\end{align}
To evaluate the second moment of the estimator after randomized data sharing is applied, we need to partition $[M]^2$ into more parts since the values of $d_{j_1}$, $d_{j_2}$, and $\mathbb{E} \left[ \rv{Z}_{j_1}^{(t)}\rv{Z}_{j_2}^{(t)}\right]$, will depend on whether the considered examples are private or not and whether they belong to the same class of not. The partitions are denoted by $\sset{J}_{\text{same}}^{\text{p-p}}$, $\sset{J}_{\text{diff}}^{\text{p-p}}$. $\sset{J}_{\text{same}}^{\text{np-p}}$, $\sset{J}_{\text{diff}}^{\text{np-p}}$, $\sset{J}_{\text{same}}^{\text{np-np}}$, and $\sset{J}_{\text{diff}}^{\text{np-np}}$. The superscript p-p indicates that both examples are private, p-np that one example is private and the other is non-private, and np-np that both examples are non-private. The subscript ``same" indicates that both examples belong to the same class and ``diff" that the two examples belong to different classes. Let $\lambda_{\text{same}}^{\text{p-p}} \triangleq \frac{\mathbb{E} \left[ \rv{Z}_{j_1}^{(t)}\rv{Z}_{j_2}^{(t)}\right]}{(1-p)^2d_{j_1}d_{j_2}}$ for $(j_1,j_2)\in \sset{J}_{\text{same}}^{\text{p-p}}$, and define similarly the quantities $\lambda_{\text{diff}}^{\text{p-p}}$, $\lambda_{\text{same}}^{\text{np-p}}$, $\lambda_{\text{diff}}^{\text{np-p}}$, $\lambda_{\text{same}}^{\text{np-np}}$, and $\lambda_{\text{diff}}^{\text{np-np}}$.  Hence, we have
\begin{multline} \label{secmY}
    \mathbb{E} \left[ \| \hat{\vvec{g}}^{(t)}_{\Rv{Y}} \|^2 \right] = \lambda_{\text{same}}^{\text{p-p}} \sum_{(j_1,j_2)\in \sset{J}_{\text{same}}^{\text{p-p}}}  \langle \vvec{g}^{(t)}_{j_1}, \vvec{g}^{(t)}_{j_2} \rangle + \lambda_{\text{diff}}^{\text{p-p}} \sum_{(j_1,j_2)\in \sset{J}_{\text{diff}}^{\text{p-p}}}  \langle \vvec{g}^{(t)}_{j_1}, \vvec{g}^{(t)}_{j_2} \rangle + \lambda_{\text{same}}^{\text{np-p}} \sum_{(j_1,j_2)\in \sset{J}_{\text{same}}^{\text{np-p}}}  \langle \vvec{g}^{(t)}_{j_1}, \vvec{g}^{(t)}_{j_2} \rangle \\ + \lambda_{\text{diff}}^{\text{np-p}} \sum_{(j_1,j_2)\in \sset{J}_{\text{diff}}^{\text{np-p}}}  \langle \vvec{g}^{(t)}_{j_1}, \vvec{g}^{(t)}_{j_2} \rangle + \lambda_{\text{same}}^{\text{np-np}} \sum_{(j_1,j_2)\in \sset{J}_{\text{same}}^{\text{np-np}}}  \langle \vvec{g}^{(t)}_{j_1}, \vvec{g}^{(t)}_{j_2} \rangle + \lambda_{\text{diff}}^{\text{np-np}} \sum_{(j_1,j_2)\in \sset{J}_{\text{diff}}^{\text{np-np}}}  \langle \vvec{g}^{(t)}_{j_1}, \vvec{g}^{(t)}_{j_2} \rangle.
\end{multline}
\begin{proposition} \label{prop1}
For $p\in[0,1)$ and $d\in \{0,1,\ldots,N-1\}$, we have
\begin{equation*}
    \lambda_{\text{same}}^{\text{p-p}}=\frac{1}{1-p},~~~ \lambda_{\text{diff}}^{\text{p-p}}=1,~~~ \lambda_{\text{same}}^{\text{np-p}}=\frac{1+d(1-p)}{(1-p)(d+1)},~~~ 1\leq \lambda_{\text{diff}}^{\text{np-p}}\leq \lambda_{\text{same}}^{\text{np-p}},~~~ 1\leq \lambda_{\text{same}}^{\text{np-np}} \leq \frac{1+d(1-p)}{(1-p)(d+1)},~~~ 1\leq \lambda_{\text{diff}}^{\text{np-np}} \leq \lambda_{\text{same}}^{\text{np-np}}.
\end{equation*}
\end{proposition} 
\noindent From Proposition~\ref{prop1}, \eqref{secmX}, and~\eqref{secmY}, we obtain
\begin{multline} \label{secmY2}
    \mathbb{E} \left[ \| \hat{\vvec{g}}^{(t)}_{\Rv{Y}} \|^2 \right] = \mathbb{E} \left[ \| \hat{\vvec{g}}^{(t)}_{\Rv{X}} \|^2 \right] + \left( \lambda_{\text{same}}^{\text{np-p}} - \frac{1}{1-p} \right) \sum_{(j_1,j_2)\in \sset{J}_{\text{same}}^{\text{np-p}}}  \langle \vvec{g}^{(t)}_{j_1}, \vvec{g}^{(t)}_{j_2} \rangle + \left( \lambda_{\text{diff}}^{\text{np-p}} -1 \right)\sum_{(j_1,j_2)\in \sset{J}_{\text{diff}}^{\text{np-p}}}  \langle \vvec{g}^{(t)}_{j_1}, \vvec{g}^{(t)}_{j_2} \rangle \\ + \left( \lambda_{\text{same}}^{\text{np-np}}-\frac{1}{1-p} \right) \sum_{(j_1,j_2)\in \sset{J}_{\text{same}}^{\text{np-np}}}  \langle \vvec{g}^{(t)}_{j_1}, \vvec{g}^{(t)}_{j_2} \rangle + \left(\lambda_{\text{diff}}^{\text{np-np}}-1 \right) \sum_{(j_1,j_2)\in \sset{J}_{\text{diff}}^{\text{np-np}}}  \langle \vvec{g}^{(t)}_{j_1}, \vvec{g}^{(t)}_{j_2} \rangle.
\end{multline}
By rearranging the terms we get
\begin{multline} \label{eqall}
    \mathbb{E} \left[ \| \hat{\vvec{g}}^{(t)}_{\Rv{X}} \|^2 - \| \hat{\vvec{g}}^{(t)}_{\Rv{Y}} \|^2 \right] =  \left(  \frac{1}{1-p} - \lambda_{\text{same}}^{\text{np-p}} \right) \sum_{(j_1,j_2)\in \sset{J}_{\text{same}}^{\text{np-p}}}  \langle \vvec{g}^{(t)}_{j_1}, \vvec{g}^{(t)}_{j_2} \rangle - \left( \lambda_{\text{diff}}^{\text{np-p}} -1 \right)\sum_{(j_1,j_2)\in \sset{J}_{\text{diff}}^{\text{np-p}}}  \langle \vvec{g}^{(t)}_{j_1}, \vvec{g}^{(t)}_{j_2} \rangle \\ + \left( \frac{1}{1-p}- \lambda_{\text{same}}^{\text{np-np}} \right) \sum_{(j_1,j_2)\in \sset{J}_{\text{same}}^{\text{np-np}}}  \langle \vvec{g}^{(t)}_{j_1}, \vvec{g}^{(t)}_{j_2} \rangle - \left(\lambda_{\text{diff}}^{\text{np-np}}-1 \right) \sum_{(j_1,j_2)\in \sset{J}_{\text{diff}}^{\text{np-np}}}  \langle \vvec{g}^{(t)}_{j_1}, \vvec{g}^{(t)}_{j_2} \rangle.
\end{multline}
\begin{assumption} \label{ass1}
We assume that
$$\sum_{(j_1,j_2)\in \sset{J}_{\text{same}}^{\text{np-p}}}  \langle \vvec{g}^{(t)}_{j_1}, \vvec{g}^{(t)}_{j_2} \rangle \geq \max \bigg\{ 0, \sum_{(j_1,j_2)\in \sset{J}_{\text{diff}}^{\text{np-p}}}  \langle \vvec{g}^{(t)}_{j_1}, \vvec{g}^{(t)}_{j_2} \rangle \bigg\}~~~\text{and}~~~ \sum_{(j_1,j_2)\in \sset{J}_{\text{same}}^{\text{np-np}}}  \langle \vvec{g}^{(t)}_{j_1}, \vvec{g}^{(t)}_{j_2} \rangle \geq \max \bigg \{ 0,  \sum_{(j_1,j_2)\in \sset{J}_{\text{diff}}^{\text{np-np}}}  \langle \vvec{g}^{(t)}_{j_1}, \vvec{g}^{(t)}_{j_2} \rangle \bigg\}.$$
\end{assumption}
\noindent  We justify this assumption based on our intuition that the scalar products of gradients computed on examples from the same class tend to have positive and higher values compared to those computed on examples from different classes, provided that the iterative algorithm is converging properly. We also substantiate this intuition with simulation results over MNIST given in Fig.~\ref{scalar_comparison}. More precisely, we expect gradients computed on examples belonging to the same class to align well in the feature space, leading to large and positive scalar products. Such alignments would indicate that the gradients are reinforcing each other's contributions towards adjusting the model parameters to better classify instances of that class. Conversely, for gradients computed on examples belonging to different classes, there might not be as much alignment between the gradients, leading to lower and potentially negative scalar products. In this case, the gradients may be pushing the decision boundaries away from each other or towards different directions in the feature space, reflecting the differences between the classes. Based on Assumption~\ref{ass1}
and Proposition~\ref{prop1} we get
\begin{align} 
    \mathbb{E} \left[ \| \hat{\vvec{g}}^{(t)}_{\Rv{X}} \|^2 - \| \hat{\vvec{g}}^{(t)}_{\Rv{Y}} \|^2 \right] &\geq \left(\frac{2-p}{1-p} - \lambda_{\text{same}}^{\text{np-p}} - \lambda_{\text{diff}}^{\text{np-p}}  \right) \sum_{(j_1,j_2)\in \sset{J}_{\text{same}}^{\text{np-p}} }  \langle \vvec{g}^{(t)}_{j_1}, \vvec{g}^{(t)}_{j_2} \rangle + \left(\frac{2-p}{1-p}- \lambda_{\text{same}}^{\text{np-np}}- \lambda_{\text{diff}}^{\text{np-np}}\right) \sum_{(j_1,j_2)\in \sset{J}_{\text{same}}^{\text{np-np}}}  \langle \vvec{g}^{(t)}_{j_1}, \vvec{g}^{(t)}_{j_2} \rangle, \\
    &\geq \frac{p}{1-p} \frac{d-1}{d+1} \sum_{(j_1,j_2)\in \sset{J}_{\text{same}}^{\text{np-p}} \cup \sset{J}_{\text{same}}^{\text{np-np}}}  \langle \vvec{g}^{(t)}_{j_1}, \vvec{g}^{(t)}_{j_2} \rangle,
\end{align}
where $\left \lvert \sset{J}_{\text{same}}^{\text{np-p}} \cup \sset{J}_{\text{same}}^{\text{np-np}}\right \rvert=\left \lvert \sset{J}_{\text{same}}\right \rvert- \left \lvert \sset{J}_{\text{same}}^{\text{p-p}}\right \rvert=MK-MK(1-c)^2=MKc(2-c)$.

\newpage
\begin{remark}
The bound derived in Theorem~\ref{thm2} is trivial for $d=1$. This is a consequence of Assumption~\ref{ass1}, which does not make any assumptions regarding the signs of the summations of scalar products of gradients computed on examples from different classes. Our simulation results in Fig.~\ref{scalar_comparison} suggest that these scalar products tend to have negative values that decrease in magnitude as the algorithm converges. Assuming these summations in~\eqref{eqall} to be negative, one could obtain
\begin{equation*}
     \mathbb{E} \left[ \| \hat{\vvec{g}}^{(t)}_{\Rv{X}} \|^2 - \| \hat{\vvec{g}}^{(t)}_{\Rv{Y}} \|^2 \right] \geq \frac{p}{1-p} \frac{d}{d+1} \sum_{(j_1,j_2)\in \sset{J}_{\text{same}}^{\text{np-p}} \cup \sset{J}_{\text{same}}^{\text{np-np}}}  \langle \vvec{g}^{(t)}_{j_1}, \vvec{g}^{(t)}_{j_2} \rangle.
\end{equation*}
\end{remark}

\end{document} 

\end{document}